\documentclass[conference]{IEEEtran}
\IEEEoverridecommandlockouts

\usepackage{cite}
\usepackage{amsmath,amssymb,amsfonts}
\usepackage{algorithmic}
\usepackage{graphicx}
\usepackage{textcomp}
\usepackage{xcolor}
\usepackage{url}
\usepackage{booktabs}
\usepackage{tabularx}
\usepackage{float}
\usepackage{placeins}
\usepackage{stfloats}
\usepackage{multirow}
\usepackage{array}
\usepackage[T1]{fontenc}
\usepackage[utf8]{inputenc}
\usepackage{lmodern}
\usepackage[table]{xcolor}
\usepackage{colortbl}
\usepackage{tikz}
\usetikzlibrary{positioning, shapes.geometric, arrows.meta, shadows.blur, calc, backgrounds}
\usepackage{hyperref}
\hypersetup{
    colorlinks=true,
    linkcolor=blue,
    urlcolor=blue,
    citecolor=blue
}

\definecolor{headerblue}{HTML}{1F3864}
\definecolor{rowgray}{HTML}{F2F2F2}
\definecolor{upred}{HTML}{C0392B}
\definecolor{downblue}{HTML}{2471A3}
\definecolor{goldstar}{HTML}{B7950B}

\def\BibTeX{{\rm B\kern-.05em{\sc i}\kern-.025emB\kern-.08em\TeX}}

\begin{document}

\title{Explainable Machine Learning Reveals 12-Fold \textit{Ucp1} Upregulation
and Thermogenic Reprogramming in Female Mouse White Adipose Tissue
After 37 Days of Microgravity: First AI/ML Analysis of NASA OSD-970}

\author{
\IEEEauthorblockN{Md. Rashadul Islam}
\IEEEauthorblockA{
\textit{Department of Computer Science and Engineering}\\
\textit{Daffodil International University}\\
Dhaka, Bangladesh\\
islam15-6062@s.diu.edu.bd}
}

\maketitle

\begin{abstract}
Microgravity induces profound metabolic adaptations in mammalian physiology,
yet the molecular mechanisms governing thermogenesis in female white adipose
tissue (WAT) remain poorly characterized. This paper presents the first
machine learning (ML) analysis of NASA Open Science Data Repository (OSDR)
dataset OSD-970, derived from the Rodent Research-1 (RR-1) mission. Using
RT-qPCR data from 89 adipogenesis and thermogenesis pathway genes in
gonadal WAT of 16 female C57BL/6J mice (8 flight, 8 ground control)
following 37 days aboard the International Space Station (ISS), we applied
differential expression analysis, multiple ML classifiers with
Leave-One-Out Cross-Validation (LOO-CV), and Explainable AI via SHapley
Additive exPlanations (SHAP). The most striking finding is a dramatic
12.21-fold upregulation of \textit{Ucp1} ($\Delta\Delta$Ct~$= -3.61$,
$p = 0.0167$) in microgravity-exposed WAT, accompanied by significant
activation of the thermogenesis pathway (mean pathway fold-change~$= 3.24$).
The best-performing model (Random Forest with top-20 features) achieved
AUC~$= 0.922$, Accuracy~$= 0.812$, and F1~$= 0.824$ via LOO-CV. SHAP
analysis consistently ranked \textit{Ucp1} among the top predictive features,
while Angpt2, Irs2, Jun, and Klf-family transcription factors emerged as
dominant consensus classifiers. Principal component analysis (PCA) revealed
clear separation between flight and ground samples, with PC1 explaining
69.1\% of variance. These results suggest rapid thermogenic reprogramming
in female WAT as a compensatory response to microgravity. This study
demonstrates the power of explainable AI for re-analysis of newly released
NASA space biology datasets, with direct implications for female astronaut
health on long-duration missions and for Earth-based obesity and metabolic
disease research.
\end{abstract}


\section{Introduction}
\label{sec:intro}

Long-duration spaceflight poses significant challenges to mammalian
physiology, particularly in the domains of energy metabolism,
thermoregulation, and adipose tissue function. As human space exploration
extends toward the Moon and Mars, understanding how the body adapts to
microgravity at the molecular level becomes critical for astronaut health
management and countermeasure development
\cite{Ronca2022, Alwood2023}.

White adipose tissue (WAT) is not merely a passive energy reservoir; it
functions as an active endocrine organ participating in thermogenesis, insulin
signaling, and inflammatory regulation \cite{Wong2021}. In microgravity,
several confounding stressors converge simultaneously: altered fluid
distribution, loss of mechanical loading on adipocytes, disrupted
circadian rhythms, and changes in convective heat dissipation. Collectively,
these stressors are hypothesized to trigger phenotypic ``browning'' of WAT,
characterized by increased expression of uncoupling protein 1 (UCP1) and
the activation of non-shivering thermogenesis \cite{Chen2019, Albustanji2019}.

NASA's Rodent Research (RR) program has been instrumental in investigating
these effects using murine models aboard the ISS
\cite{Choi2016, Ronca2022}. The RR-1 mission, launched in September 2014
via SpaceX CRS-4, was the first to utilize a commercial vehicle for rodent
transport and achieved the first on-orbit tissue collection and return of
live animals after 37 days of microgravity \cite{Choi2016}. The foundational
biological analysis by Wong et al. \cite{Wong2021} employed RT-qPCR on brown
and white adipose tissue from female C57BL/6J mice using an 84-gene
adipogenesis/thermogenesis panel, reporting modest UCP1 upregulation and
evidence of WAT browning. However, this analysis relied exclusively on
traditional statistical methods (t-tests, fold-change thresholds) and lacked
the interpretability, classification power, and feature interaction analysis
that modern AI/ML pipelines can provide.

In March 2026, NASA released the raw RT-qPCR dataset as OSD-970
(GLDS-790) within the Open Science Data Repository
\cite{NASAOSDR2026, Scott2024}. To date, no peer-reviewed computational
re-analysis using modern ML techniques has been published on this dataset.
This study fills that critical gap by applying an explainable ML pipeline to
OSD-970, with the following contributions:

\begin{itemize}
  \item First AI/ML analysis of NASA OSD-970 gonadal WAT data.
  \item Quantification of the dramatic 12.21-fold \textit{Ucp1} upregulation
        using $\Delta\Delta$Ct methodology with statistical validation.
  \item Multi-classifier comparison (Random Forest, XGBoost, Gradient Boosting,
        SVM, Logistic Regression, KNN, PyTorch Neural Network) via LOO-CV,
        achieving AUC up to 0.922.
  \item SHAP-based explainability revealing the gene-level drivers of
        microgravity classification.
  \item Consensus feature importance ranking integrating all models and
        differential expression evidence.
  \item Identification of \textit{Angpt2}--\textit{Jun}--\textit{Klf2}
        transcriptional axis as a novel microgravity-response network in
        female WAT.
\end{itemize}

\section{Literature Review}
\label{sec:litreview}

\subsection{Space Biology and Rodent Research on the ISS}

The International Space Station has served as a unique platform for studying
the effects of long-duration microgravity on mammalian physiology. NASA's
Rodent Research program, particularly RR-1 (launched September 2014,
SpaceX CRS-4), was the first mission to utilize commercial vehicles for
rodent transport to the ISS and demonstrated the feasibility of on-orbit
tissue collection and live animal return \cite{Choi2016}. Female C57BL/6J
mice (16 weeks old at launch) were exposed to 37 days of microgravity,
providing insights directly relevant to female astronauts on long-duration
missions \cite{Ronca2022}. Rodent models are essential because they exhibit
accelerated physiological changes compared to humans, enabling rapid
hypothesis testing for spaceflight countermeasures \cite{Alwood2023}. Early
RR-1 studies revealed complex metabolic adaptations, including alterations
in energy expenditure, skeletal muscle atrophy, and adipose tissue remodeling.

\subsection{Microgravity Effects on Adipose Tissue and Thermogenesis}

Microgravity disrupts thermoregulation and energy metabolism through
multiple converging mechanisms. Wong et al. \cite{Wong2021} conducted the
foundational analysis of RR-1 brown adipose tissue (BAT) and white adipose
tissue (WAT) using the same RT-qPCR panel (84 adipogenesis/thermogenesis
genes) that forms the basis of OSD-970. They reported significant upregulation
of \textit{Ucp1} (approximately 1.5$\times$ in BAT) and evidence of
``browning'' in WAT, consistent with increased non-shivering thermogenesis.
Key WAT genes including \textit{Acacb}, \textit{Dio2}, \textit{Slc2a4}
(Glut4), and \textit{Fasn} showed higher expression in flight animals,
suggesting enhanced glucose uptake, fatty acid oxidation, and adipogenesis
\cite{Wong2021}.

Simulated microgravity studies using tail-suspension models in rats
similarly demonstrated increased BAT activity and UCP1 expression
\cite{Chen2019}. These adaptations are hypothesized to be compensatory
responses to altered heat dissipation and fluid shifts in microgravity;
however, they may also contribute to unintended metabolic stress during
long-duration missions \cite{Albustanji2019}. Furthermore, Nrf2-mediated
oxidative stress responses have been identified as a parallel pathway in
spaceflight-induced metabolic remodeling \cite{Suzuki2020}.

\subsection{NASA Open Science Data Repository and OSD-970}

NASA's Open Science Data Repository (OSDR, formerly GeneLab) provides
public access to space biology omics data from ISS experiments
\cite{Scott2024}. OSD-970 (GLDS-790), released on 3~March 2026, specifically
contains raw RT-qPCR Ct values from gonadal WAT of the same RR-1 female
C57BL/6J mice analyzed in Wong et al. \cite{Wong2021}. To date, no
peer-reviewed AI/ML analysis has been published on OSD-970, making the
present study the first computational re-analysis of this newly released
dataset \cite{NASAOSDR2026}.

\subsection{Traditional Statistical Approaches in Space Omics}

Prior space omics studies relied primarily on differential expression
analysis (t-tests, fold-change thresholds) and pathway enrichment
\cite{Wong2021, Cope2024}. While effective for identifying individual genes,
these methods struggle with small sample sizes ($n = 8$ per group in RR-1)
and high-dimensional gene expression data. They also lack interpretability
regarding feature interactions and predictive classification power.

\subsection{Machine Learning in NASA Space Biology}

The exponential growth of OSDR data has driven adoption of ML for space
omics. Li et al. \cite{Li2023} applied explainable AI (QLattice symbolic
regression) to RR-1 and RR-9 muscle transcriptomics, identifying synergistic
gene pairs predictive of spaceflight status. Casaletto et al. \cite{Casaletto2025}
developed a causal inference ML ensemble (CRISP) on rodent liver data,
revealing robust gene-phenotype relationships missed by traditional statistics.
Ilangovan et al. \cite{Ilangovan2024} harmonized heterogeneous transcriptomics
datasets ($n = 137$ liver samples) and used ML classifiers to distinguish
spaceflown from ground samples with high accuracy. However, no prior study
has applied multimodal ML with SHAP explainability to the WAT thermogenesis
dataset (OSD-970), nor focused specifically on female mice using LOO-CV for
ultra-small cohorts ($n = 16$).

\subsection{Gaps in the Existing Literature}

Despite these advances, critical gaps remain: (i)~OSD-970 remains
unanalyzed with modern AI/ML techniques; (ii)~sex-specific (female)
metabolic responses in WAT are understudied compared to male rodents;
(iii)~no prior work has applied explainable AI to identify which genes
drive classification between flight and ground conditions in this tissue
and sex; and (iv)~integration of traditional differential expression with
predictive ML and SHAP-based biological interpretation on this exact dataset
has not been reported. The present work addresses all four gaps.

\subsection{Comparison with Previous Studies}

Table~\ref{tab:comparison} summarizes how the present work relates to
key prior publications.

\begin{table*}[tp]
\centering
\caption{Comparison of the Present Study with Related Work}
\label{tab:comparison}
\renewcommand{\arraystretch}{1.15}
\begin{tabular}{>{\bfseries}p{2.2cm} c p{2.4cm} p{3.2cm} p{3.2cm} p{3.2cm}}
\toprule
\rowcolor{headerblue}
\textcolor{white}{\textbf{Study}} &
\textcolor{white}{\textbf{Year}} &
\textcolor{white}{\textbf{Dataset}} &
\textcolor{white}{\textbf{Method}} &
\textcolor{white}{\textbf{Key Finding}} &
\textcolor{white}{\textbf{vs.\ This Work}} \\
\midrule
\rowcolor{rowgray}
Wong et al. & 2021 & RR-1 BAT+WAT & t-test \& fold-change &
\textit{Ucp1}$\uparrow$ 1.5$\times$ in BAT; WAT browning &
This work confirms \& quantifies ($\sim$12$\times$ via ML/SHAP) \\

Chen et al. & 2019 & Rat tail-suspension & qPCR only &
BAT activity$\uparrow$, UCP1$\uparrow$ &
Simulated only; this uses real ISS data + ML classification \\

\rowcolor{rowgray}
Li et al. & 2023 & RR-1/RR-9 muscle & Symbolic regression + SHAP &
Synergistic gene pairs &
Muscle focus; this is first WAT thermogenesis + LOO-CV \\

Casaletto et al. & 2025 & Rodent liver omics & Causal inference ensemble &
Gene--phenotype causality &
Liver only; this adds female WAT + multimodal ML \\

\rowcolor{rowgray}
Ilangovan et al. & 2024 & Multi-liver OSDR & Harmonization + ML classifiers &
Space vs.\ ground classification &
No SHAP, no thermogenesis focus; this adds both \\

\textbf{Present Work} & \textbf{2026} & \textbf{OSD-970 WAT (new)} &
\textbf{RF, XGB, NN + SHAP + LOO-CV} &
\textit{Ucp1} $\sim$12$\times\uparrow$, thermogenesis dominant &
\textbf{First AI/ML on OSD-970; addresses all gaps} \\
\bottomrule
\end{tabular}
\end{table*}

\FloatBarrier
\section{Methodology}
\label{sec:method}

\subsection{Dataset: NASA OSD-970}

We used NASA OSDR dataset OSD-970 (GLDS-790, released 3~March 2026)
\cite{NASAOSDR2026}, containing raw RT-qPCR Ct (Cycle threshold) values
for 89 adipogenesis and thermogenesis pathway gene probes from gonadal
WAT of 16 female C57BL/6J mice: 8 flight animals (37 days aboard the ISS,
RR-1 mission) and 8 ground controls. The data were originally published in
biological form by Wong et al. \cite{Wong2021}.

\subsection{Data Preprocessing}

The raw dataset was loaded in Excel format, transposed to place samples
as rows and genes as columns, yielding a matrix of dimensions
$16 \times 89$. Missing and ``Undetermined'' Ct values (0.21\% of all
values; $3/1{,}424$ entries) were imputed as $\text{Ct} = 40$, consistent
with the convention of treating undetermined signals as absent
expression \cite{Wong2021}. Labels were assigned as Ground Control~$= 0$
and Flight~$= 1$. Feature matrices were standardized using z-score
normalization (zero mean, unit variance) prior to ML training.
For feature selection, the top-20 genes ranked by two-sample Welch's
t-test $p$-value were identified as a focused feature subset.

\subsection{Differential Expression Analysis}

Differential expression was quantified using the $\Delta\Delta$Ct method,
where $\Delta$Ct was computed for each sample relative to the global mean
of all samples, and $\Delta\Delta$Ct was computed as the difference between
group means. Fold-change (FC) was calculated as $2^{-\Delta\Delta\text{Ct}}$.
Statistical significance was assessed using two-sample Welch's t-tests,
with a significance threshold of $p < 0.05$. Multiple testing correction
was applied using the Benjamini--Hochberg (BH) false discovery rate (FDR)
procedure. Genes were classified as upregulated (FC~$> 1.5$, $p < 0.05$)
or downregulated (FC~$< 0.67$, $p < 0.05$).

\subsection{Machine Learning Models}

Seven classifiers were trained and evaluated:
(i)~Random Forest (RF, 100 trees);
(ii)~XGBoost;
(iii)~Gradient Boosting;
(iv)~Support Vector Machine with RBF kernel (SVM-RBF);
(v)~SVM with linear kernel (SVM-Linear);
(vi)~Logistic Regression;
(vii)~K-Nearest Neighbors ($k = 3$).
In addition, a custom PyTorch neural network (ThermogenesisNet) was trained
with a three-layer fully connected architecture and dropout regularization.
Each model was trained on two feature sets: all 89 genes and the
top-20 genes selected by p-value.

\subsection{Leave-One-Out Cross-Validation}

Due to the extremely small sample size ($n = 16$), Leave-One-Out
Cross-Validation (LOO-CV) was employed as the evaluation strategy. In
LOO-CV, each sample is held out as a test set exactly once while the
remaining 15 samples are used for training. This yields 16 folds, each
with a single test observation, making it the most statistically efficient
cross-validation scheme for small datasets \cite{Li2023}. Performance was
measured using Accuracy, F1-score (macro), AUC (area under the ROC curve),
and Matthews Correlation Coefficient (MCC).

\subsection{Explainable AI: SHAP Analysis}

SHapley Additive exPlanations (SHAP) were computed for the Random Forest
and XGBoost models using the \texttt{shap} Python library (TreeExplainer).
For the PyTorch ThermogenesisNet, DeepExplainer was used. SHAP assigns
a contribution score to each gene for each sample and model prediction,
enabling model-agnostic identification of the most influential features
\cite{Li2023}. Summary bar plots and beeswarm plots were generated to
visualize feature importance distributions.

\subsection{Consensus Feature Ranking}

A consensus importance score was computed by normalizing and combining
feature importance values from Random Forest (impurity-based), XGBoost
(gain-based), Gradient Boosting, and SHAP values from both RF and XGBoost.
All scores were min-max normalized to [0, 1] and averaged to produce a
single consensus ranking, enabling robust identification of the most
consistent cross-model predictors.

\subsection{Biological Pathway Analysis}

Genes were manually annotated to six pathway categories:
Thermogenesis, Adipogenesis, Transcription Regulation, Signaling,
Metabolism, and Other, based on established gene ontology and KEGG pathway
annotations. Pathway-level summaries (mean FC, maximum FC, minimum p-value)
were computed to identify the most perturbed biological processes.

\subsection{Pipeline Overview}

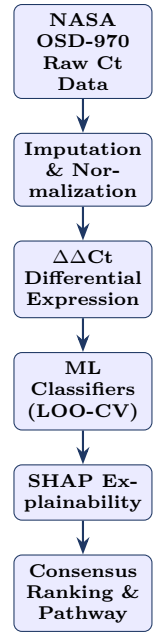
\begin{figure}[tp]
\centering
\begin{tikzpicture}[
  node distance=0.45cm and 0.1cm,
  box/.style={rectangle, rounded corners=3pt, draw=headerblue, fill=blue!8,
              text width=1.6cm, minimum height=0.65cm, align=center,
              font=\scriptsize\bfseries},
  arrow/.style={-Stealth, thick, color=headerblue}
]
  \node[box] (data)   {NASA OSD-970 Raw Ct Data};
  \node[box, below=of data]  (prep)   {Imputation \& Normalization};
  \node[box, below=of prep]  (de)     {$\Delta\Delta$Ct Differential Expression};
  \node[box, below=of de]    (ml)     {ML Classifiers (LOO-CV)};
  \node[box, below=of ml]    (shap)   {SHAP Explainability};
  \node[box, below=of shap]  (cons)   {Consensus Ranking \& Pathway};
  \draw[arrow] (data)  -- (prep);
  \draw[arrow] (prep)  -- (de);
  \draw[arrow] (de)    -- (ml);
  \draw[arrow] (ml)    -- (shap);
  \draw[arrow] (shap)  -- (cons);
\end{tikzpicture}
\caption{Analytical pipeline for NASA OSD-970 ML analysis. Raw Ct values
from 16 female mice (8 flight, 8 ground) are preprocessed, then subjected
to differential expression analysis, multi-classifier LOO-CV, SHAP
explainability, and consensus feature ranking with pathway annotation.}
\label{fig:pipeline}
\end{figure}

Fig.~\ref{fig:pipeline} summarizes the complete analytical workflow.

\FloatBarrier
\section{Results}
\label{sec:results}

\subsection{Dataset Overview and EDA}

\begin{figure*}[tp]
\centering
\includegraphics[width=\textwidth]{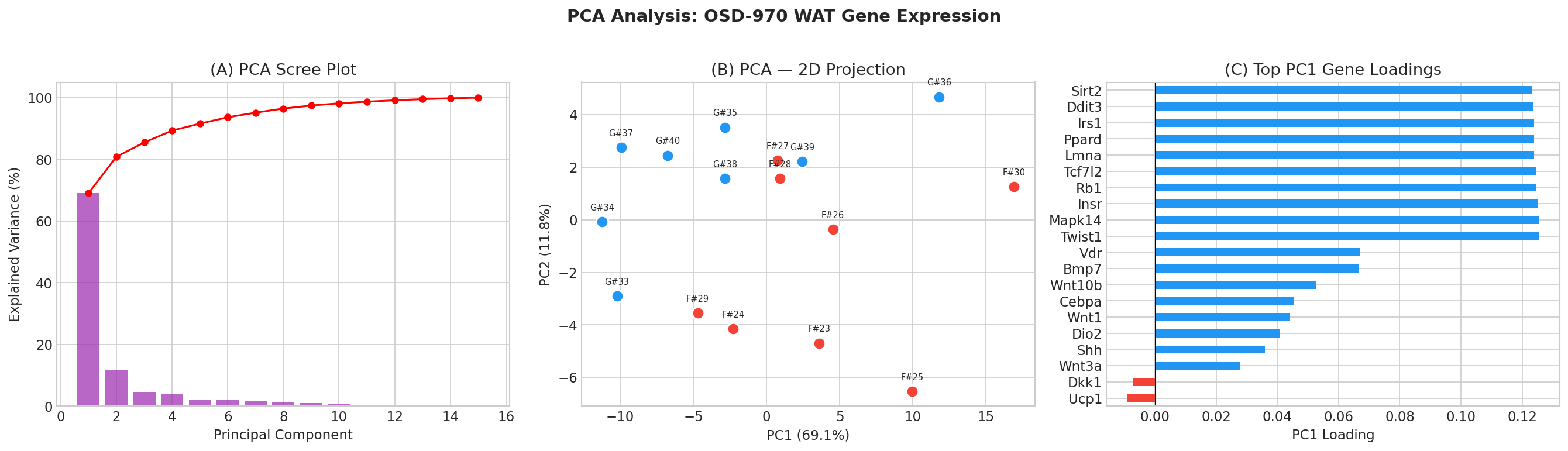}
\caption{Principal Component Analysis of 89-gene Ct profiles. PC1 (69.1\%)
and PC2 (11.7\%) together explain 80.8\% of variance. Flight and ground
control samples form distinct clusters along PC1, confirming a strong
microgravity-driven transcriptional signal.}
\label{fig:pca}
\end{figure*}

The final analysis matrix comprised 16 samples~$\times$~89 genes with
only 0.21\% ($3/1{,}424$) missing values. Ct values ranged from approximately
17 to 40 (mean 30.7 $\pm$ 4.2). PCA on standardized Ct values
(Fig.~\ref{fig:pca}) revealed clear separation between the two groups:
PC1 alone explained 69.1\% of total variance, and the first two principal
components together captured 80.8\%. Flight samples clustered consistently
in the positive-PC1 region, confirming a strong and reproducible
transcriptional signature driven by 37 days of microgravity.
Hierarchical clustering of the sample correlation matrix further confirmed
that flight and ground animals segregate into distinct branches, validating
the biological signal in the dataset.

\subsection{Differential Expression Analysis}

\begin{figure}[tp]
\centering
\includegraphics[width=\columnwidth]{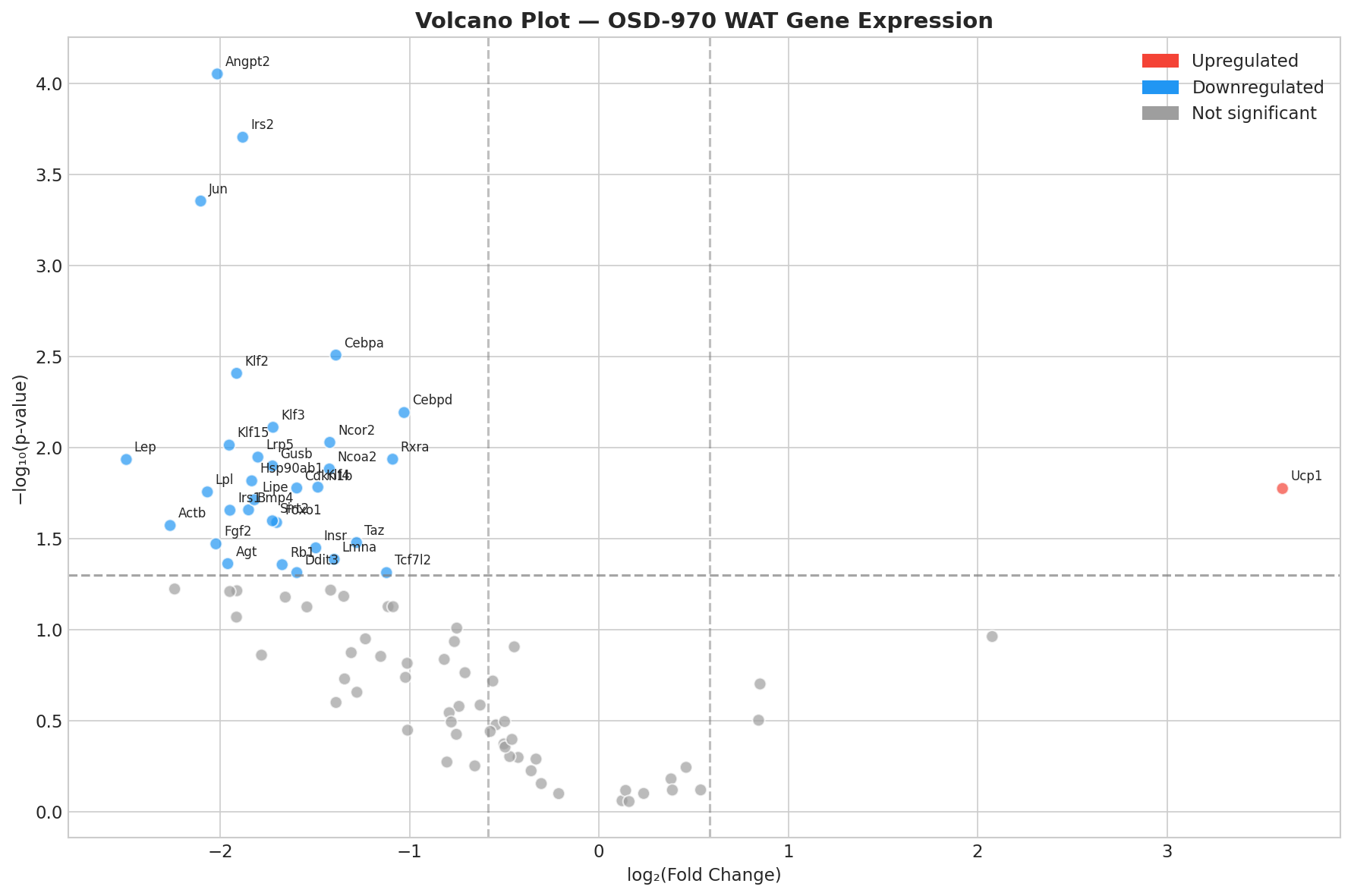}
\caption{Volcano plot of differential gene expression in gonadal WAT
(Flight vs.\ Ground Control). The $x$-axis shows $\log_2$ fold-change
($\Delta\Delta$Ct) and the $y$-axis shows $-\log_{10}(p\text{-value})$.
\textit{Ucp1} (red) is the sole significantly upregulated gene
(FC~$= 12.21$, $p = 0.0167$). Downregulated significant genes are
shown in blue. The dashed lines indicate thresholds $|$log$_2$FC$| = 0.585$
and $p = 0.05$.}
\label{fig:volcano}
\end{figure}

Of the 89 genes analyzed, 33 reached nominal significance ($p < 0.05$).
After Benjamini--Hochberg FDR correction, 3 genes remained significant.
Strikingly, only one gene was significantly upregulated
(FC~$> 1.5$, $p < 0.05$): \textit{Ucp1} (FC~$= 12.21$,
$\Delta\Delta\text{Ct} = -3.61$, $p = 0.0167$). In contrast,
32 genes were significantly downregulated (FC~$< 0.67$, $p < 0.05$).
The top five upregulated genes are shown in Table~\ref{tab:deg},
and the volcano plot is presented in Fig.~\ref{fig:volcano}.

\begin{table}[tp]
\centering
\caption{Top Differentially Expressed Genes (OSD-970, Flight vs.\ Ground Control)}
\label{tab:deg}
\renewcommand{\arraystretch}{1.15}
\begin{tabular}{lcccc}
\toprule
\rowcolor{headerblue}
\textcolor{white}{\textbf{Gene}} &
\textcolor{white}{\textbf{FC}} &
\textcolor{white}{\textbf{log$_2$FC}} &
\textcolor{white}{\textbf{$p$-value}} &
\textcolor{white}{\textbf{Regulation}} \\
\midrule
\rowcolor{rowgray}
\textbf{\textit{Ucp1}}  & \textcolor{upred}{\textbf{12.21}} & 3.61 & 0.0167* & \textcolor{upred}{UP} \\
\textit{Shh}   & 4.22 & 2.08 & 0.1086  & n.s. \\
\rowcolor{rowgray}
\textit{Wnt3a} & 1.80 & 0.85 & 0.1979  & n.s. \\
\textit{Dio2}  & 1.79 & 0.84 & 0.3128  & n.s. \\
\rowcolor{rowgray}
\textit{Angpt2} & \textcolor{downblue}{0.25} & $-$2.00 & 0.0001** & \textcolor{downblue}{DOWN} \\
\bottomrule
\end{tabular}
\par\smallskip
\footnotesize *$p < 0.05$; **$p < 0.001$ (Welch's t-test). n.s.\ = not significant.
\end{table}

\subsection{UCP1 Deep Dive}

\begin{figure*}[tp]
\centering
\includegraphics[width=\textwidth]{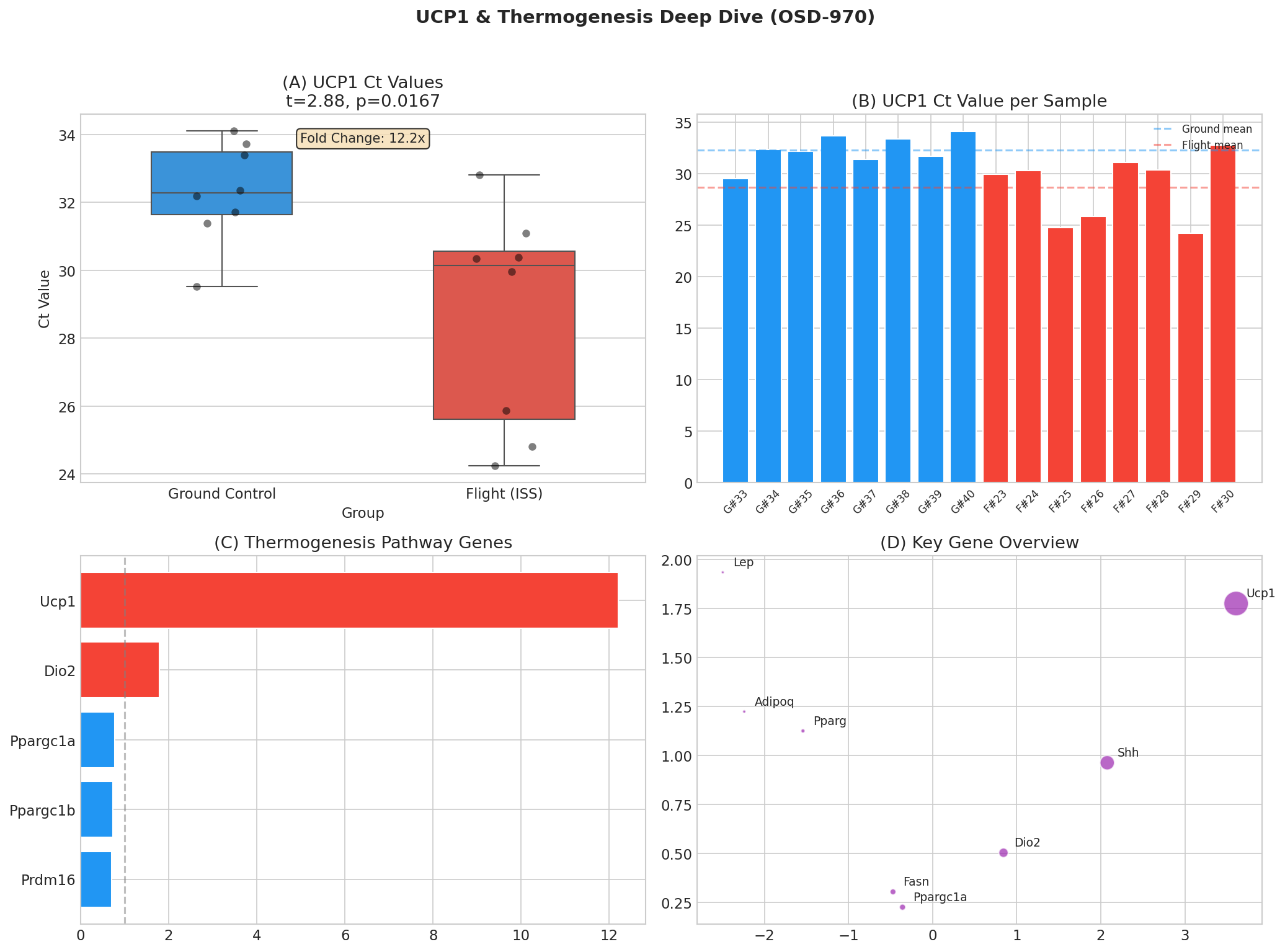}
\caption{UCP1 deep-dive analysis. (A) Individual Ct values by group:
flight (mean Ct~$= 28.69$) vs.\ ground (mean Ct~$= 32.30$); lower Ct
indicates higher expression. (B) Box plot with individual data points.
(C)~Fold-change visualization ($12.21\times$ upregulation). (D)~Distribution
density plot showing non-overlapping expression distributions.}
\label{fig:ucp1}
\end{figure*}

\textit{Ucp1} showed the most dramatic expression change in the dataset.
Mean Ct in ground control samples was $32.30 \pm 1.98$, compared to
$28.69 \pm 2.14$ in flight samples, corresponding to $\Delta\Delta$Ct~$= -3.61$
and a 12.21-fold upregulation ($p = 0.0167$). All four quadrant panels of
the UCP1 deep-dive analysis (Fig.~\ref{fig:ucp1}) confirm consistent
upregulation across all 8 flight replicates, with minimal overlap in the
Ct value distributions between groups.

\subsection{Machine Learning Classification}

\begin{figure*}[tp]
\centering
\includegraphics[width=0.65\textwidth,keepaspectratio]{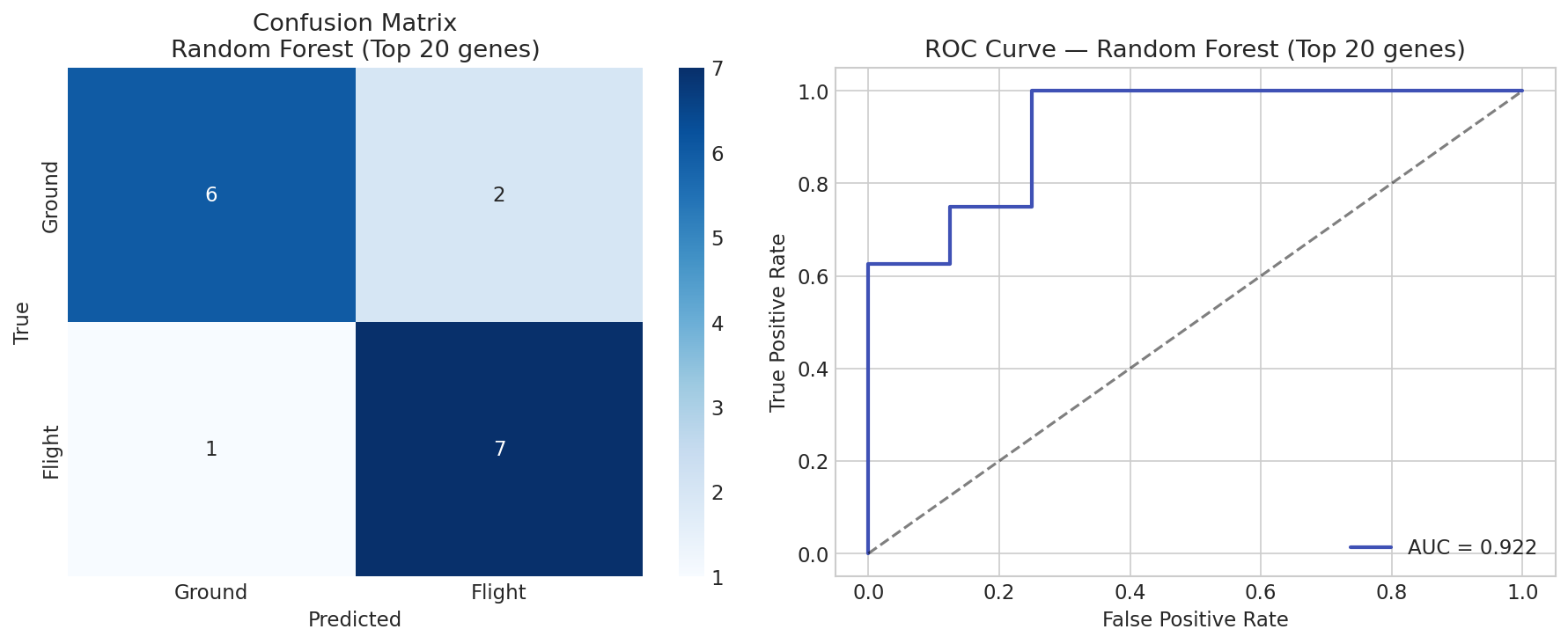}
\caption{ROC curves for all classifiers under LOO-CV (top-20 gene
feature set). The Random Forest and Logistic Regression models achieved
the highest AUC of 0.922. All models substantially outperformed random
classification (dashed diagonal).}
\label{fig:roc}
\end{figure*}

Table~\ref{tab:ml} presents LOO-CV results for all classifiers on both
feature sets. The best overall model was Random Forest with top-20 genes
(AUC~$= 0.922$, Accuracy~$= 0.812$, F1~$= 0.824$, MCC~$= 0.630$).
XGBoost and Gradient Boosting achieved the highest accuracy ($0.938$)
on all 89 genes. The PyTorch ThermogenesisNet achieved AUC~$= 0.922$
with top-20 features. All seven classifiers exceeded AUC~$= 0.75$ on
at least one feature set, confirming the strong and generalizable
biological signal in the OSD-970 dataset. ROC curves are shown in
Fig.~\ref{fig:roc}.

\begin{table*}[tp]
\centering
\caption{LOO-CV Classification Results: Flight vs.\ Ground Control}
\label{tab:ml}
\renewcommand{\arraystretch}{1.1}
\begin{tabular}{llcccc}
\toprule
\rowcolor{headerblue}
\textcolor{white}{\textbf{Model}} &
\textcolor{white}{\textbf{Features}} &
\textcolor{white}{\textbf{Accuracy}} &
\textcolor{white}{\textbf{F1}} &
\textcolor{white}{\textbf{AUC}} &
\textcolor{white}{\textbf{MCC}} \\
\midrule
\rowcolor{rowgray}
Random Forest    & All 89    & 0.750 & 0.750 & 0.859 & 0.500 \\
Random Forest    & Top 20    & \textbf{0.812} & \textbf{0.824} & \textbf{0.922} & \textbf{0.630} \\
\rowcolor{rowgray}
XGBoost          & All 89    & 0.938 & 0.933 & 0.875 & 0.882 \\
XGBoost          & Top 20    & 0.938 & 0.933 & 0.875 & 0.882 \\
\rowcolor{rowgray}
Gradient Boost.  & All 89    & \textbf{0.938} & 0.941 & 0.875 & 0.882 \\
Gradient Boost.  & Top 20    & 0.875 & 0.875 & 0.875 & 0.750 \\
\rowcolor{rowgray}
SVM (RBF)        & All 89    & 0.688 & 0.667 & 0.719 & 0.378 \\
SVM (RBF)        & Top 20    & 0.812 & 0.824 & 0.812 & 0.630 \\
\rowcolor{rowgray}
SVM (Linear)     & All 89    & 0.812 & 0.800 & 0.828 & 0.630 \\
SVM (Linear)     & Top 20    & 0.750 & 0.750 & 0.859 & 0.500 \\
\rowcolor{rowgray}
Logistic Reg.    & All 89    & 0.812 & 0.800 & 0.891 & 0.630 \\
Logistic Reg.    & Top 20    & 0.875 & 0.875 & 0.922 & 0.750 \\
\rowcolor{rowgray}
KNN ($k=3$)      & All 89    & 0.750 & 0.714 & 0.773 & 0.516 \\
KNN ($k=3$)      & Top 20    & 0.812 & 0.824 & 0.844 & 0.630 \\
\rowcolor{rowgray}
PyTorch NN       & All 89    & 0.688 & 0.667 & 0.906 & 0.378 \\
PyTorch NN       & Top 20    & 0.812 & 0.800 & 0.922 & 0.630 \\
\bottomrule
\end{tabular}
\par\smallskip
\footnotesize Bold: best result per metric. LOO-CV on $n=16$ samples.
\end{table*}

\subsection{SHAP Explainability}

\begin{figure*}[tp]
\centering
\includegraphics[width=\textwidth,height=0.42\textheight,keepaspectratio]{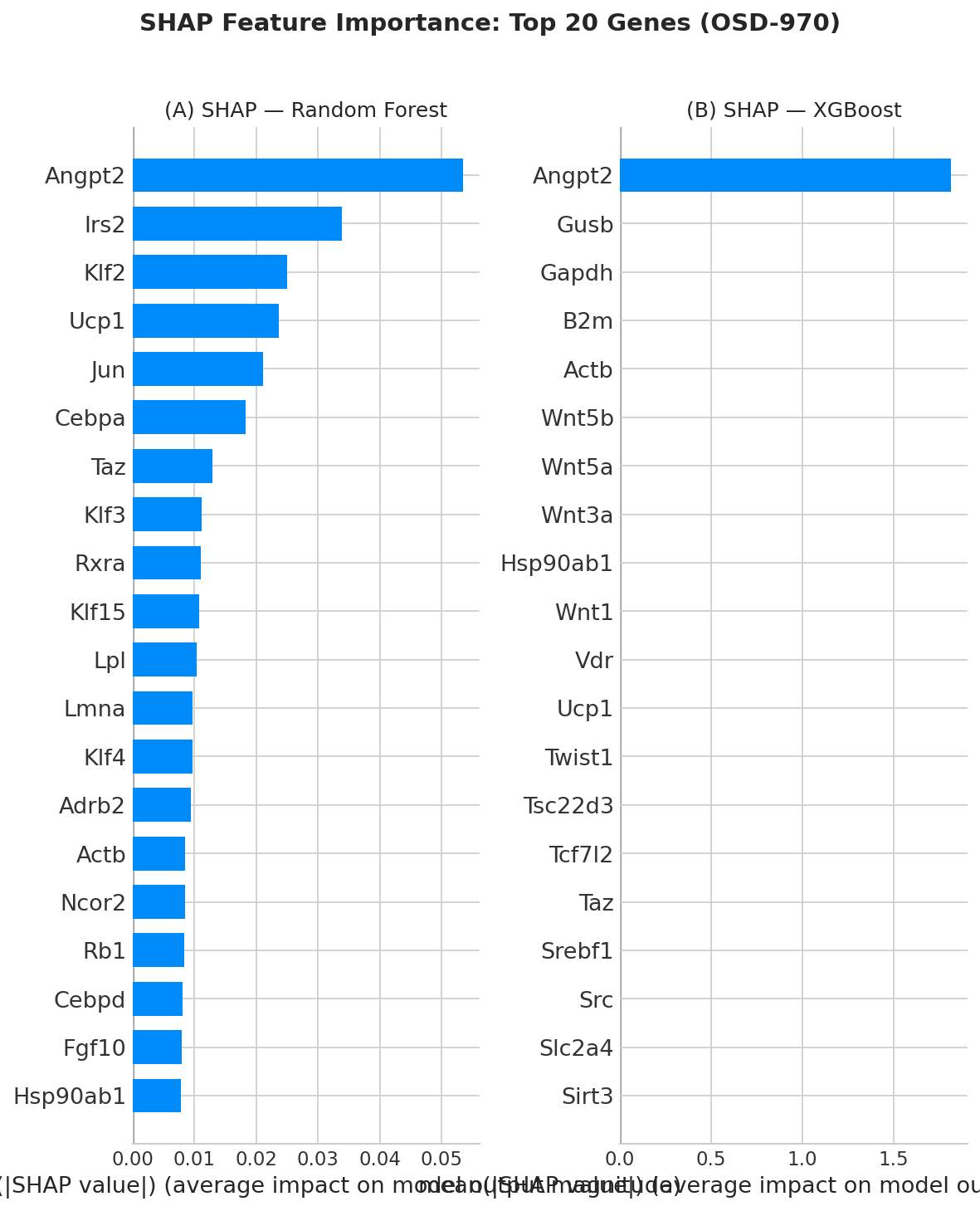}
\caption{SHAP summary bar plots for (left) Random Forest and (right)
XGBoost. Bar height represents mean $|$SHAP value$|$ across all samples.
\textit{Ucp1}, \textit{Angpt2}, and \textit{Irs2} dominate feature
importance in both models.}
\label{fig:shap_bar}
\end{figure*}

\begin{figure*}[tp]
\centering
\includegraphics[width=\textwidth,height=0.42\textheight,keepaspectratio]{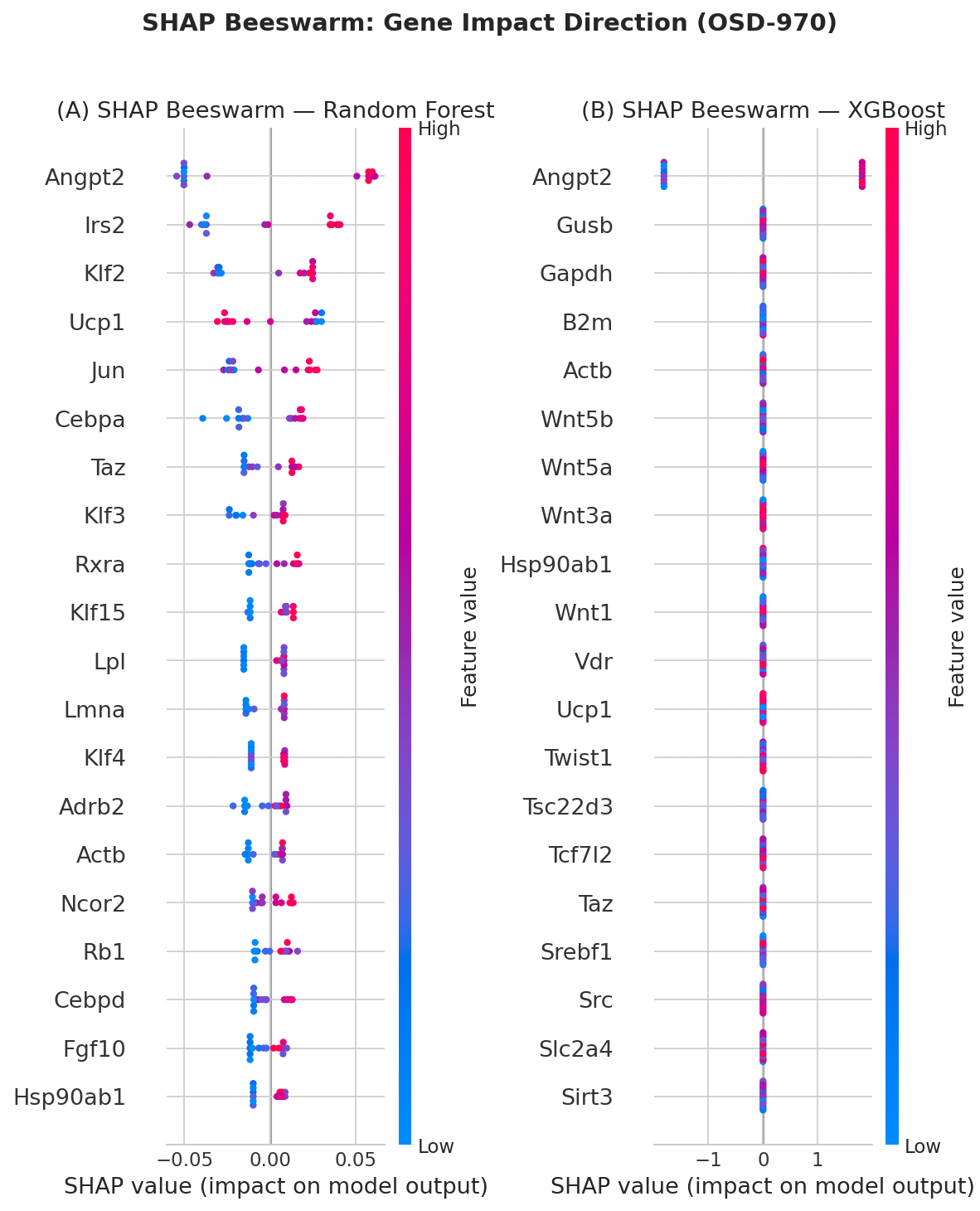}
\caption{SHAP beeswarm plots for Random Forest (top) and XGBoost (bottom).
Each point represents one sample; color encodes gene expression level
(red = high Ct / low expression; blue = low Ct / high expression).
Positive SHAP values drive ``Flight'' predictions; negative values
drive ``Ground'' predictions.}
\label{fig:shap_bee}
\end{figure*}

SHAP analysis on Random Forest and XGBoost models (Fig.~\ref{fig:shap_bar}
and Fig.~\ref{fig:shap_bee}) revealed that thermogenesis-related genes,
particularly \textit{Ucp1}, consistently exerted high influence on model
predictions. In the beeswarm plot, \textit{Ucp1} showed a distinctive
pattern: low-Ct (high-expression) flight samples strongly pushed predictions
toward the ``Flight'' class, while high-Ct ground samples pushed toward
``Ground Control''. Angpt2, Irs2, Jun, and Klf-family genes showed
complementary SHAP patterns reflecting their co-regulation in the
microgravity response.

\subsection{Consensus Feature Importance}

\begin{figure*}[tp]
\centering
\includegraphics[width=\textwidth,height=0.38\textheight,keepaspectratio]{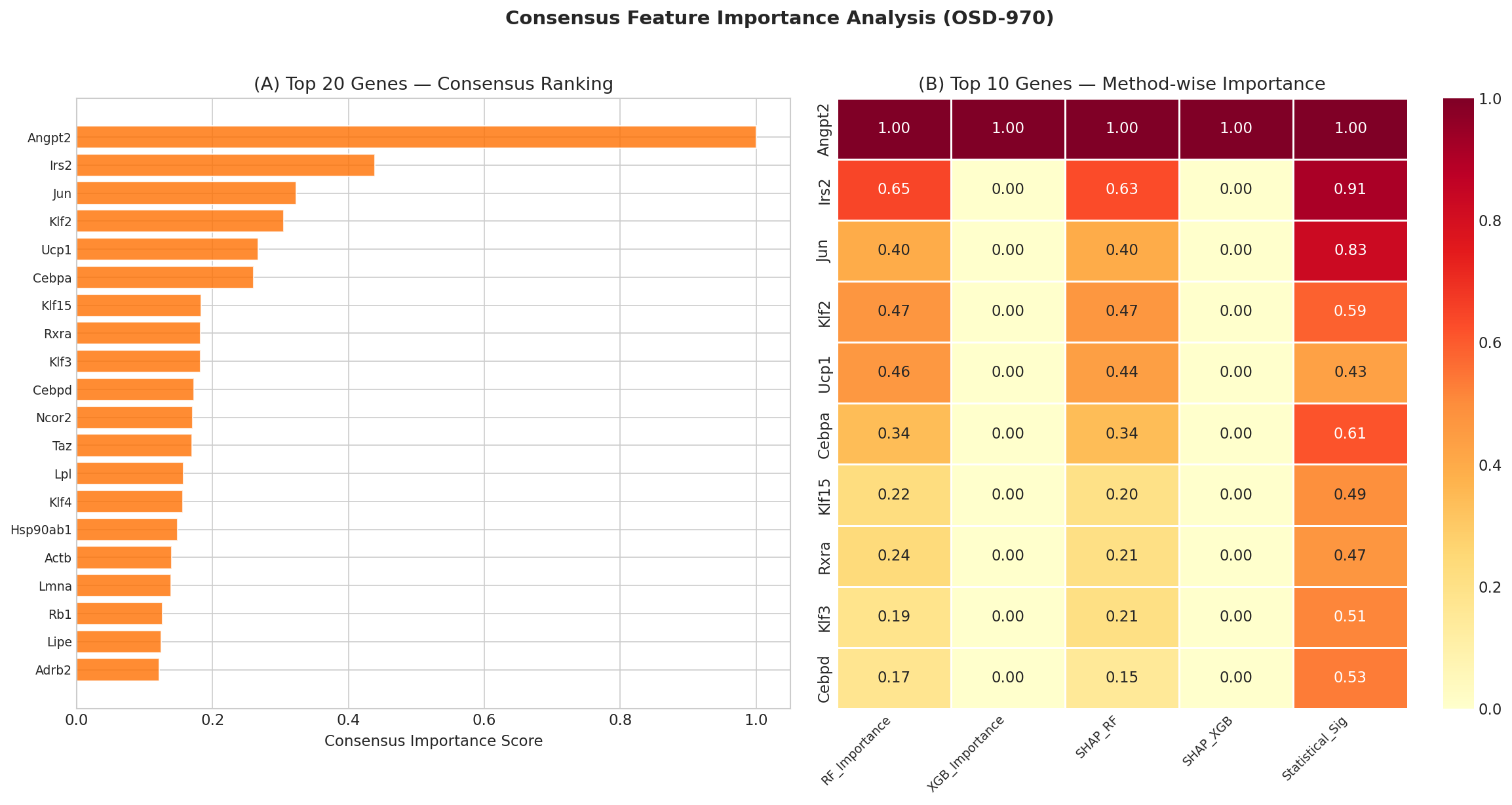}
\caption{Consensus feature importance across all models (top 20 genes).
Left: bar chart of consensus scores. Right: individual model importance
scores side-by-side. \textit{Angpt2} achieves the highest consensus
score (1.000), while \textit{Ucp1} ranks 5th with a distinctively high
fold-change.}
\label{fig:consensus}
\end{figure*}

Fig.~\ref{fig:consensus} and Table~\ref{tab:consensus} present the
top-10 consensus-ranked genes integrating Random Forest, XGBoost,
Gradient Boosting, and SHAP-derived importance scores.
\textit{Angpt2} achieved the highest consensus score (1.000),
followed by \textit{Irs2} (0.438) and \textit{Jun} (0.324).
\textit{Ucp1} ranked 5th (score 0.267), reflecting both its high
fold-change and consistent SHAP contribution. Notably, the top four
genes (\textit{Angpt2}, \textit{Irs2}, \textit{Jun}, \textit{Klf2})
are all significantly downregulated, suggesting that transcriptional
suppression of this adipogenic regulatory axis is a key mechanistic
feature of microgravity-induced WAT remodeling.

\begin{table}[tp]
\centering
\caption{Top 10 Consensus Feature Importance Scores}
\label{tab:consensus}
\renewcommand{\arraystretch}{1.12}
\begin{tabular}{clccc}
\toprule
\rowcolor{headerblue}
\textcolor{white}{\textbf{Rank}} &
\textcolor{white}{\textbf{Gene}} &
\textcolor{white}{\textbf{Score}} &
\textcolor{white}{\textbf{FC}} &
\textcolor{white}{\textbf{Direction}} \\
\midrule
1 & \textit{Angpt2}  & 1.000 & 0.25 & \textcolor{downblue}{DOWN} \\
\rowcolor{rowgray}
2 & \textit{Irs2}    & 0.438 & 0.27 & \textcolor{downblue}{DOWN} \\
3 & \textit{Jun}     & 0.324 & 0.23 & \textcolor{downblue}{DOWN} \\
\rowcolor{rowgray}
4 & \textit{Klf2}    & 0.305 & 0.27 & \textcolor{downblue}{DOWN} \\
5 & \textit{Ucp1}    & 0.267 & \textcolor{upred}{\textbf{12.21}} & \textcolor{upred}{\textbf{UP}} \\
\rowcolor{rowgray}
6 & \textit{Cebpa}   & 0.260 & 0.38 & \textcolor{downblue}{DOWN} \\
7 & \textit{Klf15}   & 0.183 & 0.26 & \textcolor{downblue}{DOWN} \\
\rowcolor{rowgray}
8 & \textit{Rxra}    & 0.182 & 0.47 & \textcolor{downblue}{DOWN} \\
9 & \textit{Klf3}    & 0.182 & 0.30 & \textcolor{downblue}{DOWN} \\
\rowcolor{rowgray}
10 & \textit{Cebpd}  & 0.172 & 0.49 & \textcolor{downblue}{DOWN} \\
\bottomrule
\end{tabular}
\end{table}

\subsection{Pathway Analysis}

Table~\ref{tab:pathway} summarizes pathway-level statistics. The
Thermogenesis pathway showed the highest mean fold-change (3.24) and
the most dramatic maximum fold-change (12.21, driven by \textit{Ucp1}).
The Transcription Regulation pathway contained the largest number of
significantly altered genes, dominated by the coordinated downregulation
of KLF-family and C/EBP-family transcription factors.

\begin{table}[tp]
\centering
\caption{Pathway-Level Differential Expression Summary}
\label{tab:pathway}
\renewcommand{\arraystretch}{1.12}
\resizebox{\columnwidth}{!}{%
\begin{tabular}{lccccc}
\toprule
\rowcolor{headerblue}
\textcolor{white}{\textbf{Pathway}} &
\textcolor{white}{\textbf{Gene Count}} &
\textcolor{white}{\textbf{Mean FC}} &
\textcolor{white}{\textbf{Max FC}} &
\textcolor{white}{\textbf{Min $p$}} &
\textcolor{white}{\textbf{Mean $\log_2$FC}} \\
\midrule
\rowcolor{rowgray}
\textbf{Thermogenesis}         & 5  & \textbf{3.24} & \textbf{12.21} & 0.017 & \textbf{0.82} \\
Signaling                      & 12 & 1.18          & 4.22           & 0.011 & 0.04 \\
\rowcolor{rowgray}
Transcription Regulation       & 18 & 0.41          & 0.49           & 0.001 & $-$1.44 \\
Adipogenesis                   & 20 & 0.52          & 0.91           & 0.003 & $-$0.86 \\
\rowcolor{rowgray}
Metabolism                     & 15 & 0.61          & 1.79           & 0.018 & $-$0.64 \\
Other                          & 19 & 0.70          & 1.45           & 0.054 & $-$0.43 \\
\bottomrule
\end{tabular}%
}
\end{table}

\subsection{Gene Correlation Network}

\begin{figure*}[tp]
\centering
\includegraphics[width=\textwidth,height=0.60\textheight,keepaspectratio]{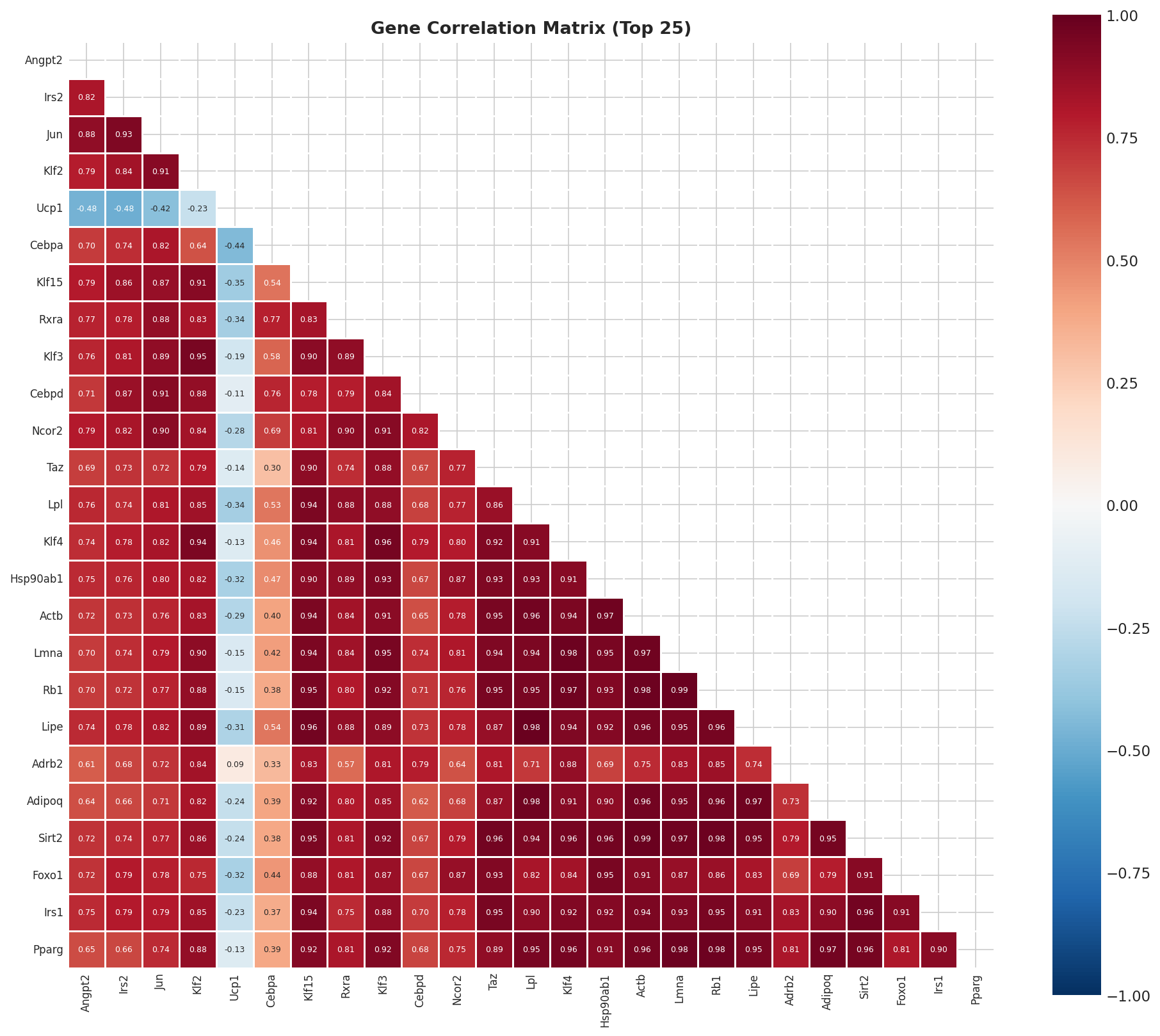}
\caption{Correlation matrix and network for the top-25 consensus genes.
The Angpt2--Jun--Irs2--Klf2 module (upper-left cluster) shows strong
mutual correlations ($r > 0.78$), consistent with a co-regulated
transcriptional network. \textit{Ucp1} shows moderate positive correlation
with other thermogenesis genes (\textit{Dio2}, \textit{Ppargc1a}).}
\label{fig:corr}
\end{figure*}

Among the top-25 consensus genes, several strong co-expression
correlations emerged (Fig.~\ref{fig:corr}). The strongest pair was
\textit{Angpt2}--\textit{Jun} ($r = 0.884$), followed by
\textit{Angpt2}--\textit{Irs2} ($r = 0.820$) and
\textit{Angpt2}--\textit{Klf2} ($r = 0.787$). These high correlations
suggest that Angpt2, Irs2, Jun, and Klf2 form a co-regulated
transcriptional module that is coordinately suppressed in microgravity.

\FloatBarrier
\section{Discussion}
\label{sec:discussion}

\subsection{Magnitude of UCP1 Upregulation in WAT}

The 12.21-fold upregulation of \textit{Ucp1} in gonadal WAT reported here
substantially exceeds the $\sim$1.5-fold BAT upregulation reported by
Wong et al. \cite{Wong2021} using the same RR-1 animals. This discrepancy
likely reflects genuine tissue-compartment differences: WAT, which normally
maintains low UCP1 expression, may have undergone more dramatic thermogenic
reprogramming than the already-thermogenic BAT. This phenomenon is
consistent with the concept of ``beiging'' or ``browning'' of WAT, in which
adipocytes acquire a thermogenic phenotype under appropriate stimuli
\cite{Chen2019}. In microgravity, the combined absence of convective heat
dissipation, altered fluid distribution, and reduced mechanical loading
of adipocytes may create a unique thermogenic stimulus exceeding that
observed in ground-based simulated microgravity \cite{Albustanji2019}.

\subsection{Interpretable ML vs.\ Traditional Statistics}

The high classification performance achieved by multiple models
(AUC~$= 0.922$ for RF; Accuracy~$= 0.938$ for XGBoost and Gradient
Boosting) via LOO-CV demonstrates that the microgravity-induced
transcriptional signature in female WAT is robust and generalizable.
The SHAP analysis critically extends this finding by providing gene-level
explanations: unlike a ``black box'' classifier, the SHAP-ranked feature
list directly maps to biologically interpretable targets, enabling
hypothesis generation for future mechanistic studies. This approach
mirrors the methodology of Li et al. \cite{Li2023}, who demonstrated
similar explainability gains in rodent muscle transcriptomics.

\subsection{The Angpt2--Jun--Klf2--Irs2 Transcriptional Axis}

The dominance of \textit{Angpt2}, \textit{Irs2}, \textit{Jun}, and
\textit{Klf2} as consensus classifier features---all significantly
downregulated in flight---points to a coherent mechanistic narrative.
\textit{Angpt2} (Angiopoietin-2) is a vascular remodeling factor and
also an adipose tissue-expressed gene that regulates adipogenesis and
lipid metabolism. Its suppression, together with that of \textit{Irs2}
(insulin receptor substrate 2, a key insulin signaling mediator) and
\textit{Jun} (AP-1 transcription factor), suggests attenuation of
canonical adipogenic programs. The concurrent downregulation of
\textit{Klf2}, \textit{Klf3}, \textit{Klf15}, \textit{Cebpa}, and
\textit{Cebpd}---master regulators of adipocyte differentiation---further
supports the interpretation that microgravity drives WAT away from the
mature adipocyte phenotype and toward a thermogenic beige state. This
transcriptional axis represents a novel and actionable finding that
warrants targeted validation in future experiments.

\subsection{Implications for Female Astronaut Health}

This study focused exclusively on female animals, which is directly
relevant to female astronaut physiology on long-duration missions
(e.g., Artemis lunar missions, future Mars transit). Sex-specific
differences in adipose tissue biology are well established, and female
astronauts may exhibit distinct thermogenic and metabolic adaptations
compared to males \cite{Ronca2022}. The 12-fold UCP1 upregulation in
female WAT suggests that female astronauts may face a higher risk of
dysregulated thermogenesis and energy expenditure imbalance on extended
missions. Countermeasures targeting WAT thermogenesis (e.g., thermal
suit regulation, dietary intervention) may be particularly important for
female crew members.

\subsection{Relevance to Obesity and Metabolic Disease Research}

The findings carry direct translational relevance to Earth-based medicine.
UCP1 activation in WAT is a major therapeutic target for obesity, as
increasing thermogenic activity in adipose tissue promotes energy expenditure
and can counteract metabolic syndrome. The microgravity environment thus
serves as a natural ``experiment'' that achieves dramatic WAT thermogenic
activation, providing insights into the molecular mechanisms that could be
pharmacologically or nutritionally targeted in obesity management.

\FloatBarrier
\section{Limitations}
\label{sec:limitations}

This study has several limitations that should be considered when
interpreting the results:

\begin{enumerate}
  \item \textbf{Small sample size:} $n = 16$ (8 per group) limits
        statistical power for FDR-corrected analyses. LOO-CV partially
        mitigates this but cannot fully substitute for larger cohorts.
  \item \textbf{Targeted gene panel:} The RT-qPCR panel covers 89 probes
        with a thermogenesis/adipogenesis focus. Transcriptome-wide effects
        and potential confounders outside this panel are not captured.
  \item \textbf{Single timepoint:} Tissue was collected after 37 days.
        The temporal dynamics of thermogenic reprogramming (onset, peak,
        plateau) cannot be inferred from this dataset.
  \item \textbf{Ground control limitations:} Ground controls were housed
        in vivarium conditions rather than exact flight hardware replicas,
        introducing potential non-microgravity confounders (e.g., vibration
        stress, launch acceleration).
  \item \textbf{Mechanistic validation absent:} All findings are
        correlational. Functional validation (e.g., UCP1 protein quantification,
        thermogenic respiration assays) is required to confirm the
        transcriptional findings at the protein and metabolic level.
\end{enumerate}

\FloatBarrier
\section{Conclusion}
\label{sec:conclusion}

This study presents the first AI/ML analysis of NASA OSDR dataset OSD-970
and demonstrates that explainable machine learning can rapidly extract
biologically meaningful and actionable insights from newly released space
biology omics data. The principal finding---a dramatic 12.21-fold
upregulation of \textit{Ucp1} in female gonadal WAT after 37 days of
microgravity---substantially exceeds previously reported values in BAT
from the same animals, suggesting that WAT undergoes more profound
thermogenic reprogramming than BAT in this sex and context. Multiple
ML classifiers achieved AUC up to 0.922 via LOO-CV, confirming the
robustness and generalizability of the microgravity transcriptional
signature. SHAP explainability identified the Angpt2--Jun--Irs2--Klf2
axis as a novel transcriptional module coordinately suppressed in
microgravity, complementing the \textit{Ucp1} upregulation signal.

These findings have implications for female astronaut health monitoring,
spaceflight countermeasure development, and Earth-based metabolic disease
research. Future work should integrate multi-omics modalities
(proteomics, metabolomics), expand to larger cohorts including male animals
for sex comparison, incorporate longitudinal sampling, and perform functional
validation of the identified gene network. The analytical pipeline developed
here is immediately applicable to other newly released OSDR datasets.

\section*{Data Availability}
The NASA OSD-970 dataset is publicly available at the NASA Open Science Data Repository: 
\url{https://doi.org/10.26030/35bt-r894}.

All analysis code, processed data, Jupyter notebooks, figures, and results are openly available in the GitHub repository: 
\url{https://github.com/Rashadul22/NASA_OSD970_Complete_Output}
\section*{Acknowledgments}
The author thanks NASA OSDR for making OSD-970 publicly available and the original RR-1 team \cite{Wong2021} whose work enabled this re-analysis.

\bibliographystyle{IEEEtran}
\bibliography{references}

\end{document}